%% file: acl2023.tex
\pdfoutput=1

\documentclass[11pt]{article}

\usepackage{ACL2023}

\usepackage{times}
\usepackage{latexsym}

\usepackage[T1]{fontenc}

\usepackage[utf8]{inputenc}

\usepackage{microtype}

\usepackage{inconsolata}

\usepackage[linesnumbered,ruled,commentsnumbered]{algorithm2e}
\usepackage{bm}
\usepackage{breqn}
\usepackage{graphicx}
\usepackage{amssymb}
\usepackage{multirow}
\usepackage{array}
\usepackage{makecell}

\newcolumntype{C}[1]{>{\centering\let\newline\\\arraybackslash\hspace{0pt}}m{#1}}

\title{Reinforcement Learning for Topic Models}

\author{Jeremy Costello \and Marek Z. Reformat \\
  Department of Electrical and Computer Engineering \\
  University of Alberta \\
  \texttt{\{jeremy1, reformat\}@ualberta.ca} \\
  }

\begin{document}
\maketitle

\begin{abstract}
    \input{ABSTRACT/abstract}
\end{abstract}

\section{Introduction}
\input{INTRODUCTION/introduction}

\section{Related Work}
\input{RELATED_WORK/related_work}

\section{Background}
\input{BACKGROUND/background}

\section{Methodology}
\input{METHODOLOGY/methodology}

\section{Results}
\input{RESULTS/results}

\section{Discussion}
\input{DISCUSSION/discussion}

\section{Conclusion}
\input{CONCLUSION/conclusion}

\section{Limitations}
\input{POST/limitations}

\section{Ethics Statement}
\input{POST/ethics.tex}

\section*{Acknolwedgements}
\input{POST/acknowledgements.tex}

\bibliography{custom}
\bibliographystyle{acl_natbib}

\appendix
\input{APPENDIX/appendix}

\end{document}

%% file: ABSTRACT/abstract.tex
We apply reinforcement learning techniques to topic modeling by replacing the variational autoencoder in ProdLDA with a continuous action space reinforcement learning policy. We train the system with a policy gradient algorithm REINFORCE. Additionally, we introduced several modifications: modernize the neural network architecture, weight the ELBO loss, use contextual embeddings, and monitor the learning process via computing topic diversity and coherence for each training step. Experiments are performed on 11 data sets. Our unsupervised model outperforms all other unsupervised models and performs on par with or better than most models using supervised labeling. Our model is outperformed on certain data sets by a model using supervised labeling and contrastive learning. We have also conducted an ablation study to provide empirical evidence of performance improvements from changes we made to ProdLDA and found that the reinforcement learning formulation boosts performance.

%% file: INTRODUCTION/introduction.tex
The internet contains large collections of unlabeled textual data. Topic modeling is a method to extract information from this text by grouping documents into topics and linking these topics with words describing them. Classical techniques for topic modeling, the most popular being Latent Dirichlet Approximation (LDA) \citep{blei2003latent}, have recently begun to be overtaken by Neural Topic Models (NTM) \citep{zhao2021topic}.

ProdLDA \citep{srivastava2017autoencoding} is a NTM using a product of experts in place of the mixture model used in classical LDA. ProdLDA uses a variational autoencoder (VAE) \citep{kingma2013auto} to learn distributions over topics and words. ProdLDA improved on NVDM \citep{miao2016neural} by explicitly approximating the Dirichlet prior from LDA with a Gaussian distribution and using the Adam optimizer \citep{kingma2014adam} with a higher momentum and learning rate.

Perceiving Reinforcement Learning (RL) as probabilistic inference has brought practices of such an inference into the RL field \citep{dayan1997using} \citep{levine2018reinforcement}. New algorithms using these techniques include MPO \citep{abdolmaleki2018maximum} and VIREL \citep{fellows2019virel}. MPO optimizes the evidence lower bound (ELBO), which is the same optimization objective used in VAEs.


Inspired by the adoption of probabilistic inference techniques in RL, we look to apply RL techniques to probabilistic inference in the realm of topic models. We use REINFORCE, the simplest policy gradient (PG) algorithm, to train a model which parameterizes a continuous action space, corresponding to the distribution of topics for each document in the topic model. We keep the product of experts from ProdLDA to compute the distribution of words for each document in the topic model.

\begin{figure*}[ht]
    \centering
    \includegraphics[width=0.775\linewidth]{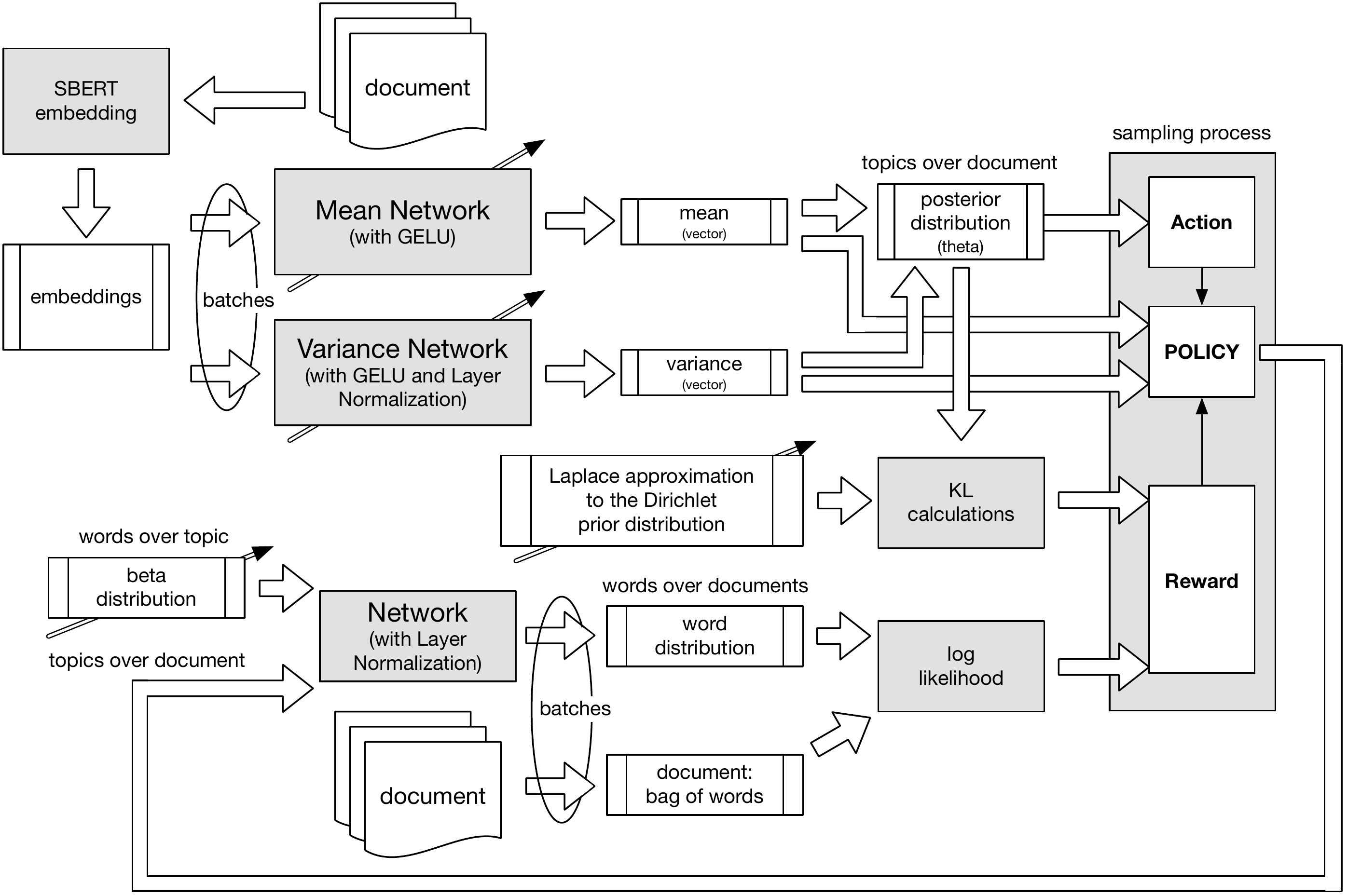}
    \caption{Architecture Diagram: gray boxes - processing; white boxes - models/data/information; arrows across boxes - tune-ability}
    \label{fig:diagram}
\end{figure*}

We additionally improve our topic model by using Sentence-BERT (SBERT) embeddings \citep{reimers-2019-sentence-bert} rather than bag-of-word (BoW) embeddings, modernizing the neural network (NN) architecture, adding a weighting term to the ELBO, and tracking topic diversity and coherence metrics throughout training. The model architecture is shown in \autoref{fig:diagram}. 
Our method outperforms most other topic models. It is beaten only
on some data sets by advanced methods using document labels for supervised learning, while our procedure is fully unsupervised.

%% file: RELATED_WORK/related_work.tex
\citet{zhao2021topic} provide a survey of NTMs. Variations of VAEs are presented which use different distributions, correlated and structured topics, pre-trained language models, incorporate meta-data, or model on short texts rather than documents. Methods other than VAEs are also used for NTMing, including autoregressive models, generative adversarial networks, and graph NNs.

\citet{doan2021benchmarking} compare ProdLDA and NVDM, along with six other NTMs and three classical topic models, in terms of held-out document and word perplexity, downstream classification, and coherence. Scholar \citep{card2017neural}, an extension of ProdLDA taking document metadata and labels into account where possible, performed best in terms of coherence. NVDM and NVCTM \citep{liu2019neural}, an extension of NVDM which additionally models the correlation between documents, performed best in terms of perplexity and downstream classification. The other NTMs were GSM \citep{miao2017discovering}, NVLDA \citep{srivastava2017autoencoding}, NSMDM \citep{lin2019sparsemax}, and NSMTM \citep{lin2019sparsemax}. The classical topic models were non-negative matrix factorization (NMF) \citep{zhao2017online}, online LDA \citep{hoffman2010online}, and Gibbs sampling LDA \citep{griffiths2004finding}.

BERTopic \citep{grootendorst2022bertopic} and Top2Vec \citep{angelov2020top2vec} use dimensionality reduction and clustering to group document embeddings from pre-trained language models into meaningful clusters. Contextualized Topic Models (CTM) \citep{bianchi2020pre} augments the BoW embeddings used in ProdLDA with SBERT \citep{reimers-2019-sentence-bert} embeddings, resulting in an improved topic model.

\citet{dieng2020topic} develop the embedded topic model (ETM) by using word embeddings to augment a variational inference algorithm for topic modeling. Their method outperforms other topic models, especially on corpora with large vocabularies containing common and very rare words. \citet{nguyen2021contrastive} augment Scholar \citep{card2017neural} with contrastive learning \citep{hadsell2006dimensionality} and outperform all topic models compared against.

\citet{gui2019neural} use RL to filter words from documents, with reward as a combination of the resulting topic model's coherence and diversity, or how few words overlap between topics. \citet{kumar2021reinforced} use REINFORCE \citep{williams1992simple}, a PG RL algorithm, to augment ProdLDA. Their model slightly outperforms ProdLDA in terms of topic coherence.

%% file: BACKGROUND/background.tex
We briefly outline topic models, RL process, KL divergence, and contextual embeddings.

\subsection{Topic Models -- Approaches}
\paragraph{Latent Dirichlet Allocation (LDA)} \citep{blei2003latent} is a three-level hierarchical Bayesian model: documents $\rightarrow$ topics $\rightarrow$ words. Each document is a mixture over latent topics, where the topic distribution $\theta$ is randomly sampled from a Dirichlet distribution. Each topic is a multinomial distribution over vocabulary words.

\paragraph{Autoencoding Variational Inference for Topic Models (AVITM)}
 \citep{srivastava2017autoencoding} is a neural topic model using a VAE to learn a Gaussian distribution over topics. VAEs use a reparameterization trick (RT) to randomly sample from the posterior distribution to remain fully differentiable. At the time, there was no known RT for Dirichlet distributions, so AVITM used a Gaussian distribution and a Laplace approximation of the Dirichlet prior.

AVITM contains two models: NVLDA and ProdLDA. NVLDA uses the mixture model from LDA to infer a distribution over vocabulary words, while ProdLDA uses a product of experts.

\paragraph{Evidence Lower Bound (ELBO)}
is the optimization objective for AVITM. ELBO optimization \citep{jordan1999introduction} simultaneously tries to maximize the log-likelihood of the topic model and minimize the forward Kullback–Leibler (KL) divergence \citep{kullback1951information} between the posterior $P$ and prior $Q$ topic distributions.
\begin{equation}
    \textnormal{ELBO} = D_{KL}(P||Q) - {log\text{-}likelihood}
    \label{eqn:elbo}
\end{equation}

\subsection{Topic Models -- Evaluation}
\paragraph{Topic Coherence}
is a metric for evaluating topic models. It uses co-occurence in a reference corpus to measure semantic similarity between the top-K words in a topic. Topic model coherence is the average of each topic's coherence.

Normalized pointwise mutual information (NPMI) \citep{aletras2013evaluating} was the coherence measure found to correlate best with human judgment \citep{lau2014machine}. 
When computing NPMI, a window size of 20 for co-occurrence counts is used in \citet{srivastava2017autoencoding}, while \citet{dieng2020topic} uses full document co-occurrence.

NPMI coherence is calculated for each of the top-K words in a topic and averaged to obtain the coherence for that topic. The overall 
\textit{topic-coherence} is the average of the coherence for each topic. For a word $i$, the NPMI coherence is calculated according to \autoref{eqn:npmicoherence}.

\begin{equation}
    \textnormal{NPMI}(w_i) = \sum_{j}^{K-1} \frac{\log{\frac{P(w_i,w_j)}{P(w_i)P(w_j)}}}
    {-\log{P(w_i,w_j)}}
    \label{eqn:npmicoherence}
\end{equation}
where $P(w_i)$ is the probability of word $i$ occurring in a document in the corpus, and $P(w_i, w_j)$ is the probability of words $i$ and $j$ co-occurring in a document in the corpus.

\paragraph{Topic Diversity}
is another metric for evaluating topic models. It measures the uniqueness of the top-K words across all topics. \citet{dieng2020topic} use $K = 25$ for reporting topic diversity.
\begin{equation}
    {topic\text{-}diversity} = \frac{{number\text{-}of\text{-}unique\text{-}words}}{K * {number\text{-}of\text{-}topics}}
    \label{eqn:diversity}
\end{equation}

\paragraph{Topic Quality}
is a topic modeling metric introduced by \citet{dieng2020topic}.
\begin{dmath}
    {topic\text{-}quality} = 
    {topic\text{-}coherence} * {topic\text{-}diversity}
    \label{eqn:quality}
\end{dmath}

\subsection{Reinforcement Learning}
RL is a sequential decision-making framework 
focused on finding the best sequence of actions executed by an agent.
\citep{sutton2018reinforcement}. An agent takes actions $a \in \mathbf{A}$ to traverse between states $s \in \mathbf{S}$ in an environment, receiving a reward $r$ on each transition. The goal of an RL task is to find the best set of actions \textemdash referred to as the policy \textemdash which maximizes the reward. RL problems can be episodic, where the agent completes the environment and is reset, or continuing, where the agent continuously traverses the environment without reset. Through traversing the environment, the agent learns a policy $\pi$ of which actions in each state will maximize return. Return is the cumulative reward received by the agent in an episode or its lifetime. It is usually discounted by a factor $ \gamma $ to favor near-term reward over long-term reward. An alternative to discounting is the average reward formulation.

\paragraph{Policy Gradient (PG) Algorithms} Many RL algorithms learn a value function -- representing values associated with selecting specific actions -- and a corresponding policy that chooses the action or subsequent state with maximum value. PG algorithms \citep{sutton1999policy} provide an alternative approach directly learning a parameterized policy. The parameters of the policy function are optimized through stochastic gradient ascent.

\paragraph{REINFORCE}
is a Monte Carlo PG algorithm for episodic problems \citep{williams1992simple}. See \autoref{alg:reinforce}, where $\bm{\rho}$
is a vector of optimized parameters.

\begin{algorithm}
\SetKwInput{Input}{Input}
\SetKwInput{Params}{Algorithm Parameters}
\DontPrintSemicolon
\caption{REINFORCE}\label{alg:reinforce}
\Input{A differentiable parameterized policy function $\pi(a|s,\bm{\rho})$}
\Params{\\ \hspace{1.1cm}step size $\alpha>0$, 
\\ \hspace{1cm} discount factor $\gamma<1$}

Initialize $\bm{\rho}$ (e.g. $\bm{\rho} \sim N(0,0.02)$)

\For{each episode}{
    Generate an episode \\ $s_{0}$, $a_{0}$, $r_{1}$, $\dots$, 
    $s_{T-1}$, $a_{T-1}$, $r_{T}$ \\ following policy $\pi$
    
    \For{each step in the episode ($t$ from $0$ to $T - 1$)}{
        $G \leftarrow \sum_{k=t+1}^{T} \gamma ^{k-t-1} r_{k}$
        
        $\bm{\rho} \leftarrow \bm{\rho} + \alpha \gamma ^{t} G \nabla \ln{\pi(a_{t}|s_{t},\bm{\rho})}$
    }
}
\end{algorithm}

\paragraph{Continuous Action Spaces}
are one advantage of PG algorithms \citep{sutton2018reinforcement}. Parameterized policies allow action spaces that are parameterized by a probability distribution, such as a Gaussian. For Gaussian action spaces, the mean $\mu$ and standard deviation $\sigma$ are given by function approximators parameterized by $\bm{\rho}$. For a state $s$, an action $a$ is sampled from the distribution and the policy is updated according to \autoref{eqn:policy}.
\begin{equation}
    \pi(a|s,\bm{\rho}) \doteq \frac{1}{\sigma(s,\bm{\rho})\sqrt{2\pi}}\exp {\left(-\frac{(a-\mu(s,\bm{\rho}))^{2}}{2\sigma(s,\bm{\rho}^{2})}\right)}
    \label{eqn:policy}
\end{equation}

\paragraph{Kullback-Leibler (KL) Divergence} \citep{kullback1951information} measures the similarity between two probability distributions $P$ and $Q$. It is used in AVITM \citep{srivastava2017autoencoding} to force the posterior distribution parameterized by the VAE to be the Laplace approximation of the Dirichlet prior. The KL divergence calculation for $N$ topics is shown in \hyperref[eqn:kldivergence]{Equation 6}.
\begin{dmath}
    D_{KL}(P||Q) = \frac{1}{2} \sum_1^N \left(
    \frac{(\mu_P - \mu_Q)^2}{\sigma_Q^2} + \\
    \null\hfill \frac{\sigma_P^2}{\sigma_Q^2} -
    \log \frac{\sigma_P^2}{\sigma_Q^2} - 1 \right)
    \label{eqn:kldivergence}
\end{dmath}

\subsection{Contextual Embeddings}
Contextual embeddings dominate NLP tasks, replacing earlier methods, including Word2Vec \citep{mikolov2013efficient}, GloVe \citep{pennington2014glove}, and BoW. Words and sequences of words are encoded into vector embeddings by large Transformer models \citep{vaswani2017attention}.

The BoW document representation used in ProdLDA is augmented with contextual embeddings from SBERT \citet{bianchi2020pre}. They test three models: one with BoW, one with contextual embeddings, and one with both. They find that using both embeddings produces the best results, and the other two methods perform almost as well. One advantage of using solely contextual embeddings is that multilingual language models can encode documents from different languages into the same embedding space, enabling easy creation of multilingual topic models \citep{bianchi2020cross}.

\paragraph{Sentence-BERT}
is an extension of BERT using a Siamese network to extract semantically meaningful sentence embeddings \citep{reimers-2019-sentence-bert}. In contrast to BERT, this allows SBERT embeddings to be compared using dot product or cosine similarity, making SBERT more suitable for tasks such as semantic similarity search and clustering.

%% file: METHODOLOGY/methodology.tex
\subsection{Modernizing ProdLDA}
Following \citet{liu2022convnet}, we contemporize the architecture of the inference network within ProdLDA. We replace the SoftPlus activation function \citep{glorot2011deep} with a GELU activation function \citep{hendrycks2016gaussian}, replace batch normalization \citep{ioffe2015batch} with layer normalization \citep{ba2016layer}, and replace all Xavier initialization \citep{glorot2010understanding} with $\bm{\rho} \sim N(0, 0.02)$.

For the inference network, we increase the number of units in each layer from 100 to 128, add weight decay of 0.01 to each layer, and place dropout layers \citep{srivastava2014dropout} after each fully connected layer.

We replace the softmax activation after the topic distribution with an RL policy formulation (\autoref{eqn:policy}). We use a training batch size of 1024. We clip all gradients to a maximum norm of 1.0 to prevent gradient explosion \citep{pascanu2013difficulty}. Following \citet{bianchi2020pre}, we set both distributional priors as trainable parameters. We lower optimizer learning rate to 3e-4 and momentum to 0.9.

\subsection{Document Embeddings}
Following \citet{bianchi2020pre}, we replace the BoW used by ProdLDA with contextualized embeddings from SBERT. We use the "all-MiniLM-L6-v2" model for encoding unpreprocessed documents as embedding vectors. BoW embeddings, 
used to calculate the log-likelihood of the topic model, are created using preprocessed documents.

\subsection{Single-step REINFORCE with a Continuous Action Space}
We adopt the view of RL as a statistical inference method \citep{levine2018reinforcement}. The modernized inference network from ProdLDA is used to parameterize a continuous action space from which an action is sampled, and the policy is computed according to \autoref{eqn:policy}. The topic model distribution over vocabulary words uses the product of experts from ProdLDA. We use REINFORCE to train the network, with a weighted version of ELBO as the reward. Each document embedding is a state in the environment, and each episode terminates after a single step (i.e., action). Each action is a sample from the topic distribution.

\subsection{Weighted Evidence Lower Bound}
Following \citet{higgins2016beta}, we allow modifiable relative entropy between the prior and posterior by weighting the KL divergence term in the ELBO. We define a hyperparameter $\lambda$ as a multiplier on the KL divergence term.
\begin{equation}
    \text{ELBO}_{weighted} = \lambda D_{KL}(P||Q) - {log\text{-}likelihood}
    \label{eqn:elbo_weighted}
\end{equation}

%% file: RESULTS/results.tex
\subsection{Initial Experiments} \label{sec:initial}
\begin{figure}[ht!]
    \centering
    \includegraphics[width=0.9\linewidth]{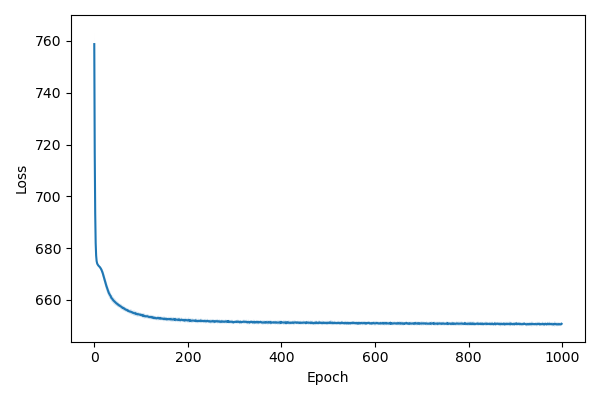}
    \caption{Loss (30 seeds): 20 Newsgroups}
    \label{fig:loss}
\end{figure}

\begin{figure}[ht!]
    \centering
    \includegraphics[width=0.9\linewidth]{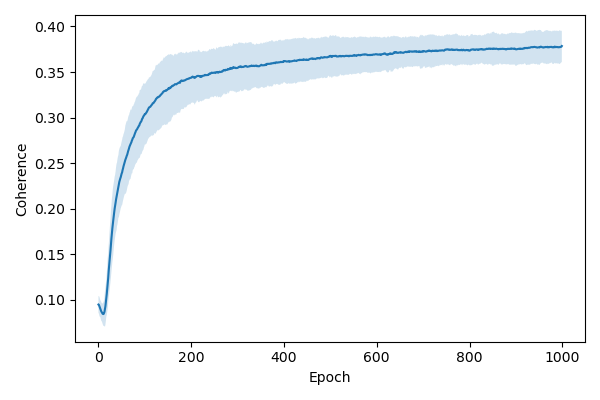}
    \caption{Topic Coherence (30 seeds): 20 Newsgroups}
    \label{fig:coherence}
\end{figure}

\begin{figure}[ht!]
    \centering
    \includegraphics[width=0.9\linewidth]{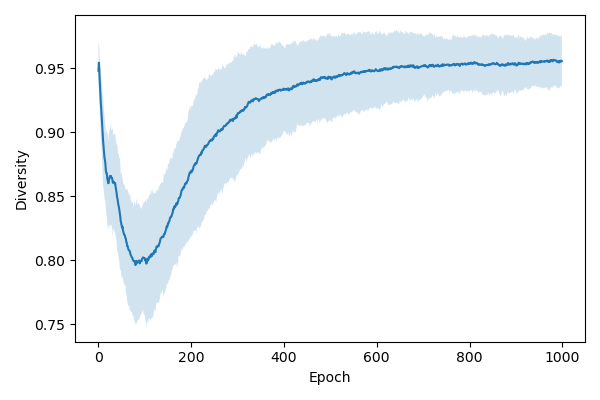}
    \caption{Topic Diversity (30 seeds): 20 Newsgroups}
    \label{fig:diversity}
\end{figure}

We initially evaluate our topic model on the 20 Newsgroups data set with 20 topics. Results averaged over 30 random seeds are shown: loss in \autoref{fig:loss}, topic coherence in \autoref{fig:coherence}, and topic diversity in \autoref{fig:diversity}. Mean and 90\% confidence intervals are plotted. Topic diversity and coherence are calculated with $K = 10$. Documents are preprocessed following \citet{bianchi2020pre} with the additional step of removing all words with less than three letters. Models are trained for 1000 epochs with the AdamW optimizer ($\alpha = 3e-4$, $\beta_1 = 0.9$, $\beta_2 = 0.999$). We use $\lambda = 5$, inference network dropout of 0.2, and no dropout after the RL policy (policy dropout). All other experiments use these same settings unless otherwise noted.

\begin{figure*}[ht]
    \centering
    \includegraphics[width=0.875\linewidth]{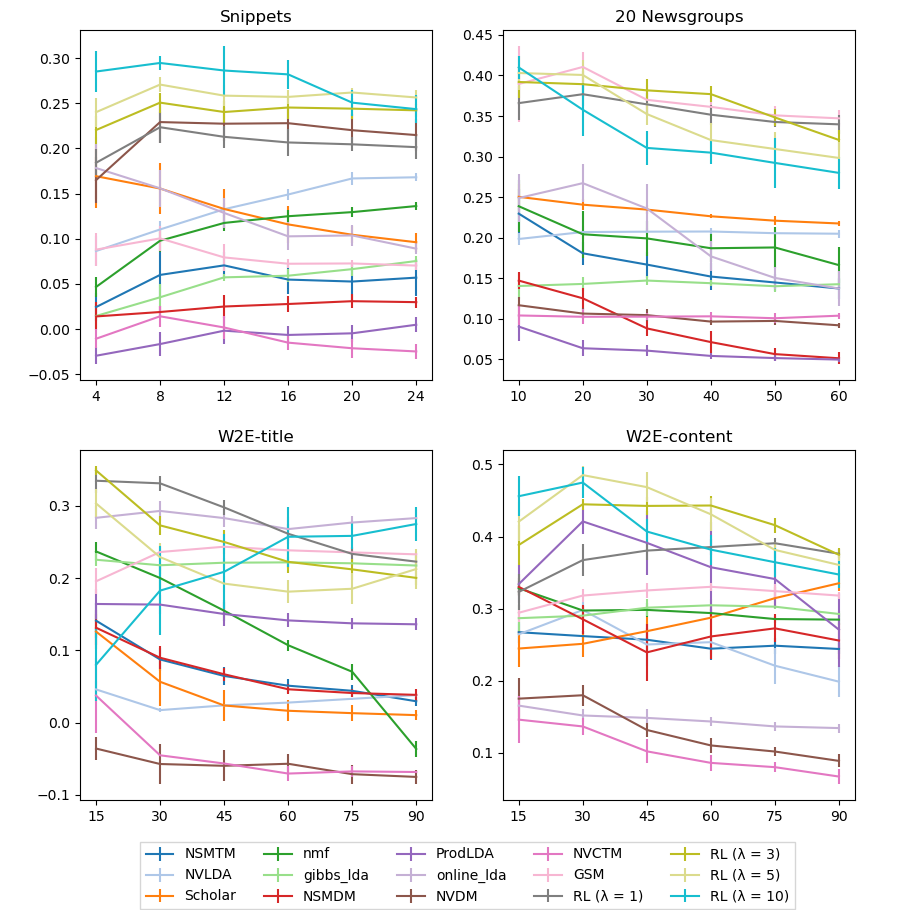}
    \caption{Comparison of RL model (ours) to BNTM models}
    \label{fig:bntmaes}
\end{figure*}

\subsection{Comparison to Other Topic Models}
We compare our method to recent topic models found in the literature.

\subsubsection{Benchmarking Neural Topic Models (BNTM)}
In the beginning, our approach is compared with all models evaluated by \citet{doan2021benchmarking}. We use their preprocessed documents and replicate their results using $K=10$ to calculate topic coherence. Following the authors, we sweep from 0.5*N topics to 3*N topics in intervals of 0.5*N (N being the "correct" number of topics for each data set). Next, we do a hyperparameter sweep over $\lambda$ of 1, 3, 5, and 10. Results are averaged over ten random seeds and shown in \autoref{fig:bntmaes}.

\subsubsection{Topic Modeling in Embedding Spaces}
Next, the comparison is done with \citet{dieng2020topic} on the New York Times data set with 300 topics and without using stop words. Results are shown in \autoref{tab:embedding}. We increase batch size to 32768 and only train for 20 epochs on one random seed. Additionally, we increase the number of units in each layer of the inference network to 512, increase dropout in the inference network to 0.5, and decrease $\lambda$ to 1. Topic diversity is calculated using $K = 25$.

\begin{table}[ht]
    \centering
    \begin{tabular}{ |C{0.2\linewidth}|c|c|c| }
        \hline
        Model & Coherence & Diversity & Quality \\
        \Xhline{2\arrayrulewidth}
        ETM & 0.18 & 0.22 & 0.0405 \\
        \hline
        RL model (ours) & $\textbf{0.24}$ & $\textbf{0.32}$ & $\textbf{0.0778}$ \\
        \hline
    \end{tabular}
    \caption{Comparison on no stop words data}
    \label{tab:embedding}
\end{table}

\subsubsection{Pre-training is a Hot Topic (PTHT)}
We also compare our model, using all metrics, with the best model as evaluated by \citet{bianchi2020pre}. Results are shown in \autoref{tab:piaht_average}. Metrics are averaged over 25, 50, 75, 100, and 150 topics: 30 seeds for each number of topics. We use the same preprocessing as the authors. We use $\lambda = 1$.

\begin{table*}[ht]
    \centering
    \begin{tabular}{ |c|c|c|c|c| }
        \hline
        Data Set & Paper & NPMI & Word2Vec & Inverse RBO \\
        \Xhline{2\arrayrulewidth}
        Wiki20K & PTHT best & 0.1823 & 0.2110 & $\textbf{0.9950}$ \\
         & RL model (ours) & $\textbf{0.2509}$ & $\textbf{0.2368}$ & 0.9799 \\
        \hline
        StackOverflow & PTHT best & 0.0280 & 0.1598 & $\textbf{0.9914}$ \\
         & RL model (ours) & $\textbf{0.1249}$ & $\textbf{0.1617}$ & 0.9860 \\
        \hline
        Google News & PTHT best & 0.1207 & 0.1325 & $\textbf{0.9965}$ \\
         & RL model (ours) & $\textbf{0.3563}$ & $\textbf{0.1485}$ & 0.9934 \\
        \hline
        Tweets2011 & PTHT best & 0.1008 & $\textbf{0.1493}$ & 0.9956 \\
         & RL model (ours) & $\textbf{0.3559}$ & 0.1417 & $\textbf{0.9962}$ \\
        \hline
        20 Newsgroups & PTHT best & 0.1300 & $\textbf{0.2539}$ & 0.9931 \\
         & RL model (ours) & $\textbf{0.2696}$ & 0.1798 & $\textbf{0.9932}$ \\
        \hline
    \end{tabular}
    \caption{Average metrics from best PTHT model (per metric) and our RL model}
    \label{tab:piaht_average}
\end{table*}

\begin{table*}[ht]
    \centering
    \begin{tabular}{ |c|c|c|c|c|c|c| }
        \hline
        Experiment & Layer Size & Inference Dropout & Policy Dropout & $\lambda$ \\
        \Xhline{2\arrayrulewidth}
        Hyperparameter Search & \{128, 512\} & \{0.2, 0.5\} & \{0.0, 0.25, 0.5\} & \{1, 5\} \\
        \hline
        20 Newsgroups & 128 & 0.5 & 0.5 & 1 \\
        IMDb Movie Reviews & 512 & 0.5 & 0.25 & 1 \\
        Wikitext-103 & 512 & 0.5 & 0.25 & 5 \\
        \hline
    \end{tabular}
    \caption{Hyperparameter search and best results per data set for RL model}
    \label{tab:hypsearch}
\end{table*}

\subsubsection{Contrastive Learning for NTM (CLNTM)}
We compare results with the contrastive Scholar model from \citet{nguyen2021contrastive}. For each data set we perform a hyperparameter search with 50 topics. Search ranges and best results for each data set are shown in \autoref{tab:hypsearch}. We use the best hyperparameters from this search for final training runs with 50 and 200 topics. We train for 2000 epochs. Results are averaged over 30 random seeds and shown in \autoref{tab:contrastive}.

\begin{figure}[ht!]
    \centering
    \includegraphics[width=0.9\linewidth]{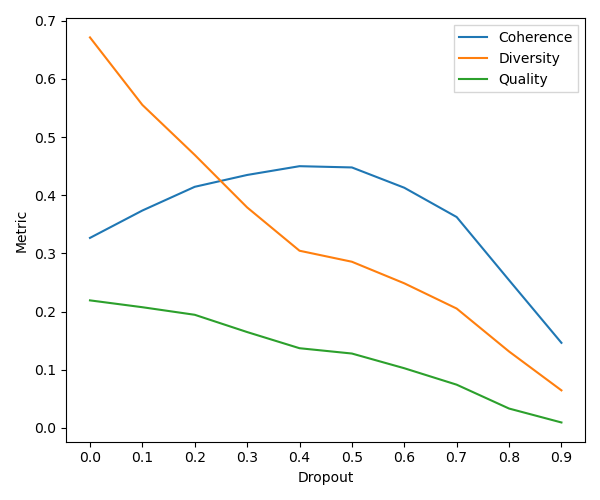}
    \caption{Dropout sweep for 20 Newsgroups}
    \label{fig:dropout}
\end{figure}

To show the tradeoff between topic diversity and coherence, we perform a sweep over policy dropout from 0 to 0.9 at intervals of 0.1 using the 20 Newsgroups data set with 50 topics. Other hyperparameters are kept the same. We train for 2000 epochs. Results are averaged over 30 random seeds and shown in \autoref{fig:dropout}.

\begin{table*}[ht]
    \centering
    \begin{tabular}{ |c|c|c|c|c|c|c| }
        \hline
        \multirow{2}{*}{Model}& \multicolumn{2}{c|}{20 Newsgroups} & \multicolumn{2}{c|}{IMDb Movie Reviews} & \multicolumn{2}{c|}{Wikitext-103} \\
        \cline{2-7}
        & 50 Topics & 200 Topics & 50 Topics & 200 Topics & 50 Topics & 200 Topics \\
        \Xhline{2\arrayrulewidth}
        Contrastive Scholar & 0.334 & 0.280 & 0.197 & $\textbf{0.188}$ & $\textbf{0.497}$ & $\textbf{0.478}$ \\
        RL model (ours) & $\textbf{0.449}$ & $\textbf{0.308}$ & $\textbf{0.199}$ & 0.139 & 0.432 & 0.268 \\
        \hline
    \end{tabular}
    \caption{Comparison to CLNTM}
    \label{tab:contrastive}
\end{table*}

\subsection{Ablation Study}
To provide empirical evidence that performance improvements come from the RL policy formulation, we do a study ablating relevant changes from the final RL model down to the original ProdLDA model. All comparisons are performed on the 20 Newsgroups data set with 20 topics and use the same settings as \autoref{sec:initial}. Results are averaged over 30 random seeds and shown in \autoref{tab:ablation}.

\begin{table*}[ht]
    \centering
    \begin{tabular}{ |c|c|C{0.05\textwidth}|c|c|c|c| }
        \hline
        RL Policy & Embedding & $\lambda$ & $\theta$ Softmax & $\theta$ / Policy Dropout & Coherence & Diversity \\
        \Xhline{2\arrayrulewidth}
        $\checkmark$ & SBERT & 5 & $\times$ & 0.0 & $\textbf{0.3848}$ & $\textbf{0.9530}$ \\
        $\times$ & SBERT & 5 & $\times$ & 0.0 & 0.2795 & 0.453 \\
        $\checkmark$ & BoW & 5 & $\times$ & 0.0 & 0.3379 & 0.9403 \\
        $\checkmark$ & SBERT & 1 & $\times$ & 0.0 & 0.3414 & 0.9070 \\
        $\checkmark$ & SBERT & 5 & $\checkmark$ & 0.0 & 0.1932 & 0.6927 \\
        $\checkmark$ & SBERT & 5 & $\times$ & 0.2 & 0.3769 & 0.7315 \\
        $\times$ & BoW & 1 & $\checkmark$ & 0.2 & 0.2650 & 0.7390 \\
        \hline
    \end{tabular}
    \caption{Highlighted results from ablation study}
    \label{tab:ablation}
\end{table*}

%% file: DISCUSSION/discussion.tex
For the initial experiments on the 20 Newsgroups data set, the average loss (\autoref{fig:loss}) reaches a near plateau around the 200th epoch. Past this epoch, coherence (\autoref{fig:coherence}) continues to increase slowly, and topic diversity (\autoref{fig:diversity}) increases substantially until around the 400th epoch, past which it also continues to increase slowly. It shows that training beyond a plateau in loss can still improve NTM performance.

Compared to \citet{doan2021benchmarking}, the RL model performs on par with or better than other models across all four data sets, while the performance of other models varies greatly between data sets. On the Snippets, 20 Newsgroups, and W2E-content data sets, the RL model with lower values of $\lambda$ usually performs better as the number of topics increases. However, it reverses on the W2E-title data set where $\lambda = 10$ outperforms $\lambda = 1$ on the two highest number of topics.

The RL model outperforms the Labeled ETM model from \citet{dieng2020topic} in topic diversity, coherence, and quality. Furthermore, this comparison had no pruning of stop words, showing the RL model can deal with vocabularies containing many common words.

Compared to \citet{bianchi2020pre}, the RL model significantly outperforms all other models on all data sets evaluated in terms of NPMI coherence. Furthermore, the RL model performs similarly to the best of the other models in terms of inverse RBO. We state the topic diversity used by \citet{dieng2020topic} is a more useful metric than inverse RBO, as it usually has a higher variance in values and is more intuitive to understand. For Word2Vec coherence, the RL model performs on par with the best of the other models, except when compared to ETM \citep{dieng2020topic} on the 20 Newsgroups data set.

If we consider models from \citet{nguyen2021contrastive}, our RL model performs similarly on 50 topics but worse on 200 topics. The RL model's performance on larger topic sizes and vocabularies could be improved by adding supervised labels, applying contrastive learning, scaling up inference layer sizes, or performing a hyperparameter sweep with 200 topics.

Topic diversity and coherence values should be provided when reporting topic model performance. In \autoref{fig:dropout}, the highest topic quality is achieved when there is no policy dropout. Topic diversity can be sacrificed for some gain in coherence. Applications of topic models may want to maximize topic diversity, coherence, or both. The description of topic model performance should reflect this.

In the ablation study, removing the RL policy formulation causes the model to perform worse than the original one. 
It confirms RL policy augments the improvements from other changes to the model. Performance suffers the most when the softmax distribution is re-added to the topic distribution during training. To recapture the softmax distribution of topics, it can be applied to the topic distribution during inference. Adding policy dropout significantly reduces topic diversity and leads to a slight coherence reduction. Performance improves with SBERT embeddings, and the model can still reconstruct the BoW within the ELBO without direct access. Increasing $\lambda$ to 5 improves performance, but as seen from other experiments, this is only sometimes the case.

%% file: CONCLUSION/conclusion.tex
Inspired by the introduction of probabilistic inference techniques to RL, we take the approach to develop a NTM augmented with RL. Our model builds on the ProdLDA model, which uses a product of experts instead of the mixture model used in classical LDA. We improve ProdLDA by adding SBERT embeddings, an RL policy formulation, a weighted ELBO loss, and the improved NN architecture. In addition, we track topic diversity and coherence during a training process rather than only evaluating these metrics for the final model. Our fully unsupervised RL model outperforms most other topic models. It is only topped by contrastive Scholar \textemdash a method using supervised labels during training \textemdash in a few select cases.

%% file: POST/limitations.tex
The main limitation identified for our RL model is decreased performance as the vocabulary size increases. Our RL model also has a higher variance than some other topic models to which we compared. While our RL model performed well on all the data sets tested, this performance may not generalize to different data sets. The insights from the policy dropout sweep conducted may not apply to other topic models. The performance difference for NPMI coherence compared with \citet{bianchi2020pre} may be overstated since the model in that paper used a deprecated SBERT model that produces sentence embeddings of low quality\footnote{https://huggingface.co/sentence-transformers/stsb-roberta-large}. For the comparison to \citet{nguyen2021contrastive}, we used slightly different preprocessing than the authors. While the model can work on any languages with associated embedding models, all data sets used in this paper were in English. Our model has additional hyperparameters compared to some other models. So, it may require more tuning and, therefore, more GPU computing. The initial model was developed on a system with 8GB of RAM and a Nvidia GTX 1060 with 3GB of VRAM for a total of approximately 100 GPU hours. A single run of the model for 1000 epochs on this GPU requires less than an hour. Experiments using the New York Times data set were run on a system with 256GB of RAM and a Nvidia RTX 3090 for approximately 100 GPU hours. All other experiments were run on a system with 128GB of RAM and a Nvidia TITAN RTX for approximately 600 GPU hours.

%% file: POST/ethics.tex
All data sets used in this paper are cited. The New York Times data set\footnote{https://catalog.ldc.upenn.edu/LDC2008T19} is licensed under "The New York Times Annotated Corpus Agreement"\footnote{https://catalog.ldc.upenn.edu/license/the-new-york-times-annotated-corpus-ldc2008t19.pdf}. The Tweets2011 corpus\footnote{https://trec.nist.gov/data/tweets/} is available under the "TREC 2011 Microblog Dataset Usage Agreement"\footnote{https://trec.nist.gov/data/tweets/tweets2011-agreement.pdf} which additionally requires following the "Twitter terms of service"\footnote{https://twitter.com/en/tos}. All other data sets are obtained from the recent literature. No sensitive information is used or inferred in this paper. The risk of harm from our model is low. Any artifacts in this paper are used following their intended use cases.

%% file: POST/acknowledgements.tex
We would like to thank Federico Bianchi for assistance in finding data sets. We would like to thank the creators and maintainers of Python and the following Python packages: lda, torch, numpy, biterm, scipy, gensim, tqdm, transformers, nltk, sentence\_transformers, sklearn, and pandas. We would like to thank the following GitHub users for code inspiration: maifeng, smutahoang, dice-group, shion-h, karpathy, estebandito22, akashgit, and MilaNLProc.

%% file: APPENDIX/appendix.tex
\section{KL Divergence in RL}
\input{APPENDIX/kl_divergence}

\section{Data Sets}
\input{APPENDIX/datasets}

\section{Evaluation Metrics}
\input{APPENDIX/evaluation}

\section{Expanded Results}
\input{APPENDIX/further_results}

\section{Model Parameter Count}
\input{APPENDIX/parameters}

\section{Future Work}
\input{APPENDIX/future_work}

\input{APPENDIX/final_tables}

%% file: APPENDIX/kl_divergence.tex
\begin{table*}[ht]
    \centering
    \begin{tabular}{ |c|c|c|c|c| }
        \hline
        Data Set & Comparison Paper & Training Docs & Test Docs & Vocab Size \\
        \Xhline{2\arrayrulewidth}
        \multirow{4}{*}{20 Newsgroups} & This one & \multirow{3}{*}{11,314} & \multirow{3}{*}{7,532} & \multirow{3}{*}{2,000} \\
        & \citep{bianchi2020pre} & & & \\
        & \citep{nguyen2021contrastive} & & & \\
        \cline{2-5}
        & \citep{doan2021benchmarking} & 15,465 & N/A & 4,134 \\
        \hline
        New York Times & \citep{dieng2020topic} & 1,864,470 & N/A & 10,283 \\
        \hline
        Snippets & \citep{doan2021benchmarking} & 12,295 & N/A & 4,666 \\
        \hline
        W2E-title & \citep{doan2021benchmarking} & 105,457 & N/A & 3,703 \\
        \hline
        W2E-content & \citep{doan2021benchmarking} & 83,548 & N/A & 10,508 \\
        \hline
        Wiki20K & \citep{bianchi2020pre} & 20,000 & N/A & 2,000 \\
        \hline
        StackOverflow & \citep{bianchi2020pre} & 16,407 & N/A & 2,236 \\
        \hline
        Google News & \citep{bianchi2020pre} & 11,108 & N/A & 8,099 \\
        \hline
        Tweets2011 & \citep{bianchi2020pre} & 2,472 & N/A & 5,097 \\
        \hline
        IMDb Movie Reviews & \citep{nguyen2021contrastive} & 25,000 & 25,000 & 5,000 \\
        \hline
        Wikitext-103 & \citep{nguyen2021contrastive} & 28,472 & 60 & 20,000 \\
        \hline
    \end{tabular}
    \caption{Data Sets - Documents and Vocabularies}
    \label{tab:dataset_docs_vocab}
\end{table*}

KL divergence has recently become popular in continuous action space RL algorithms. One application is to prevent policy updates from making large changes to the policy that could result in poorer performance. Two algorithms using KL divergence for this are TRPO \citep{schulman2015trust} and MPO \citep{abdolmaleki2018maximum}. Another application is for optimistic RL \citep{filippi2010optimism} \citep{kobayashi2022optimistic}. \citet{vieillard2020leverage} investigate the usage of KL divergence as regularization in RL. KL divergence has also been used in optimal control \citep{kappen2012optimal}, which is closely related to RL.

%% file: APPENDIX/datasets.tex
We evaluate models on the test set where available, and on the training set if there is no test set. Coherence and diversity for the training and test set are the same, as they are evaluated on the word distribution over topics which doesn't change per document. In the code, training coherence and diversity are computed after each batch, while test coherence and diversity are computed after each epoch. Number of training/test documents and vocabulary sizes are shown in \autoref{tab:dataset_docs_vocab}. Average original and preprocessed training document lengths are shown in \autoref{tab:dataset_doc_length}.

\begin{table*}[ht]
    \centering
    \begin{tabular}{ |c|c|C{3cm}|C{3cm}| }
        \hline
        \multirow{2}{*}{Data Set} & \multirow{2}{*}{Comparison Paper} & \multicolumn{2}{c|}{Average Training Document Length} \\
        \cline{3-4}
        & & Original & Preprocessed \\
        \Xhline{2\arrayrulewidth}
        \multirow{4}{*}{20 Newsgroups} & This one & 287.5 & 95.9 \\
        \cline{2-4}
        & \citep{bianchi2020pre} & \multirow{2}{*}{287.5}& \multirow{2}{*}{107.6} \\
        & \citep{nguyen2021contrastive} & & \\
        \cline{2-4}
        & \citep{doan2021benchmarking} & N/A & 73.5 \\
        \hline
        New York Times & \citep{dieng2020topic} & 558.1 & 484.5 \\
        \hline
        Snippets & \citep{doan2021benchmarking} & N/A & 14.4 \\
        \hline
        W2E-title & \citep{doan2021benchmarking} & N/A & 6.8 \\
        \hline
        W2E-content & \citep{doan2021benchmarking} & N/A & 209.1 \\
        \hline
        Wiki20K & \citep{bianchi2020pre} & 49.8 & 17.5 \\
        \hline
        StackOverflow & \citep{bianchi2020pre} & N/A & 4.9 \\
        \hline
        Google News & \citep{bianchi2020pre} & N/A & 6.2 \\
        \hline
        Tweets2011 & \citep{bianchi2020pre} & N/A & 8.6 \\
        \hline
        IMDb Movie Reviews & \citep{nguyen2021contrastive} & 233.8 & 101.7 \\
        \hline
        Wikitext-103 & \citep{nguyen2021contrastive} & 295.8 & 133.2 \\
        \hline
    \end{tabular}
    \caption{Data Sets - Training Document Lengths}
    \label{tab:dataset_doc_length}
\end{table*}

\subsection{20 Newsgroups}
The 20 Newsgroups data set \citep{lang1995newsweeder} consists of around 19,000 newsgroup posts from 20 topics. We perform experiments on this data set with three different preprocessing methods. For our initial experiments, we follow the preprocessing in \citet{bianchi2020pre} and additionally remove all words with less than 3 letters. For the comparisons with \citet{bianchi2020pre} and \citet{nguyen2021contrastive}, we follow the preprocessing in \citet{bianchi2020pre}. For the comparison with \citet{doan2021benchmarking}, we use their already preprocessed data set.

\subsection{New York Times}
The New York Times data set \citep{sandhaus2008new} consists of over 1.8 million articles written by the New York Times between 1987 and 2007. We follow the preprocessing from \citet{bianchi2020pre}, but do not remove stopwords.

\subsection{Snippets}
The Web Snippets data set \citep{ueda2002parametric} consists of around 12,000 snippets of text from websites linked on "yahoo.com". The snippets are grouped into 8 domains. We use the already preprocessed data set from \citet{doan2021benchmarking}.

\subsection{W2E}
The W2E data set \citep{hoang2018w2e} consists of news articles from media channels around the world. The W2E-title subset is the titles from the news articles, while the W2E-content subset is the text content of the articles. The articles are grouped into 30 topics. We use the already preprocessed data set from \citet{doan2021benchmarking}.

\subsection{Wiki20K}
The Wiki20K data set \citep{bianchi2020cross} consists of 20,000 English Wikipedia abstracts randomly sampled from DBpedia. We follow the preprocessing from \citet{bianchi2020pre}.

\subsection{StackOverflow}
The StackOverflow data set \citep{qiang2020short} consists of around 16,000 question titles randomly sampled from 20 different tags in a larger data set crawled from the website "stackoverflow.com" between July and August 2012. We use the already preprocessed data set from \citet{qiang2020short}.

\subsection{Google News}
The Google News data set \citep{qiang2020short} consists of around 11,000 titles and short samples from Google News articles clustered into 152 groups. We use the already preprocessed data set from \citet{qiang2020short}.

\subsection{Tweets2011}
The Tweets2011 data set \citep{qiang2020short} consists of around 2,500 tweets in 89 clusters sampled from the larger Tweets2011 corpus \citep{mccreadie2012building} crawled from Twitter between January and February 2011. We use the already preprocessed data set from \citet{qiang2020short}.

\subsection{IMDb Movie Reviews}
The IMDb Movie Reviews data set \citep{maas2011learning} consists of 50,000 movie reviews, each with an associated sentiment label, from the website "imdb.com". We follow the preprocessing from \citet{bianchi2020pre}.

\subsection{Wikitext-103}
The Wikitext-103 data set \citep{merity2016pointer} consists of around 28,500 Wikipedia articles classified as either Featured articles or Good articles by Wikipedia editors. We follow the preprocessing from \citet{bianchi2020pre}.

%% file: APPENDIX/evaluation.tex
\begin{table*}[ht]
    \centering
    \begin{tabular}{ |c|c| }
        \hline
        Topic Words \\
        \Xhline{2\arrayrulewidth}
        max giz bhj chz pts buf air det pit bos \\
        morality objective cramer moral livesey optilink keith homosexual clayton gay \\
        window xterm widget lib windows font usr mouse motif application \\
        gun guns militia firearms weapons cops weapon amendment semi arms \\
        team players hockey game season nhl games play teams leafs \\
        max giz bhj sale chz shipping offer monitor copies condition \\
        jesus god bible christ christians faith church christian heaven lord \\
        geb banks msg patients gordon pitt disease pain doctor medical \\
        fbi batf koresh compound atf waco sandvik udel fire kent \\
        car insurance cars dealer oil saturn honda engine bmw miles \\
        jpeg image bits display gif file program files format color \\
        clipper encryption key chip escrow keys privacy crypto secure nsa \\
        wire ground circuit connected cable atheism electrical universe keyboard output \\
        israel israeli arab jews arabs peace palestinian attacks bony villages \\
        turkish armenian armenians armenia turks serdar argic turkey genocide soviet \\
        pub ftp anonymous tar graphics privacy mailing archive motif faq \\
        moon space lunar orbit nasa spacecraft henry launch shuttle solar \\
        dog bike dod riding ride motorcycle rider bmw went cops \\
        scsi ide drive controller drives bus disk floppy bios isa \\
        stephanopoulos president jobs myers russia russian administration package launch clinton \\
        \hline
    \end{tabular}
    \caption{Initial Experiment Topic Words}
    \label{tab:initial_topic_words}
\end{table*}

We track topic diversity, coherence, perplexity, and loss for the training and test sets if applicable. Topic diversity and coherence are calculated based on the top-$K$ words in each topic, with $K$ noted for each experiment. We use NPMI coherence with co-occurence based on full document windows.

Most previous NTMs have only reported the coherence of the final model, presumably because coherence is not tracked during training for computational reasons. To enable tracking of coherence during training, we modify a vectorized implementation of UMass coherence\footnote{https://github.com/maifeng/Examples_UMass-Coherence} to calculate NPMI coherence and add caching for further speed-up. We also implement a GPU-optimized algorithm to calculate topic diversity during training.

Tracking these metrics during training provides two main benefits. The first benefit is that if training is going poorly, it can be terminated. Poor training could be caused by component collapse (low topic diversity), or if the model is unable to fit to coherent topics (low coherence). The second benefit is enabling deeper performance comparisons between models and between training runs for a single model. Most existing NTMs only track loss and perplexity during training, so additionally tracking topic diversity and coherence could provide additional insights on model performance.

%% file: APPENDIX/further_results.tex
\subsection{Topic Words from Initial Experiments}
We choose one example of the top 10 words for all 20 topics from the initial experiments on the 20 Newsgroups data set. We choose the seed with the 15th highest coherence (out of 30 seeds). Topic words are shown in \autoref{tab:initial_topic_words}. Each document in the Twenty Newsgroups data set is labeled as belonging to one of 20 categories. These 20 categories are shown in \autoref{tab:20news_categories}.

\begin{table}[ht!]
    \centering
    \begin{tabular}{ |c|c| }
        \hline
        Category \\
        \Xhline{2\arrayrulewidth}
        alt.atheism \\
        comp.graphics \\
        comp.os.ms-windows.misc \\
        comp.sys.ibm.pc.hardware \\
        comp.sys.mac.hardware \\
        comp.windows.x \\
        misc.forsale \\
        rec.autos \\
        rec.motorcycles \\
        rec.sport.baseball \\
        rec.sport.hockey \\
        sci.crypt \\
        sci.electronics \\
        sci.med \\
        sci.space \\
        soc.religion.christian \\
        talk.politics.guns \\
        talk.politics.mideast \\
        talk.politics.misc \\
        talk.religion.misc \\
        \hline
    \end{tabular}
    \caption{20 Newsgroups Categories}
    \label{tab:20news_categories}
\end{table}

\subsection{Pre-training is a Hot Topic}
We show a further comparison between the contextual embedding model from \citet{bianchi2020pre} and our RL model in \autoref{tab:piaht_npmi}. Average NPMI coherence over 30 seeds is compared for each number of topics: 25, 50, 75, 100, and 150.

\begin{table*}[ht]
\centering
  \begin{tabular}{ |c|C{0.1\linewidth}|c|c|c|c|c| }
    \hline
    \multirow{2}{*}{Data Set} & \multirow{2}{*}{Paper} & \multicolumn{5}{c|}{NPMI Coherence} \\
    \cline{3-7}
    & & 25 Topics & 50 Topics & 75 Topics & 100 Topics & 150 Topics \\
    \Xhline{2\arrayrulewidth}
    Wiki20K & PTHT & 0.17 & 0.19 & 0.18 & 0.19 & 0.17 \\
    \cline{2-7}
     & RL model (ours) & 0.33 & 0.30 & 0.25 & 0.22 & 0.19 \\
    \hline
    StackOverflow & PTHT & 0.05 & 0.03 & 0.02 & 0.02 & 0.02 \\
    \cline{2-7}
     & RL model (ours) & 0.17 & 0.14 & 0.12 & 0.11 & 0.10 \\
    \hline
    Google News & PTHT & 0.03 & 0.10 & 0.15 & 0.18 & 0.19 \\
    \cline{2-7}
     & RL model (ours) & 0.38 & 0.41 & 0.38 & 0.34 & 0.30 \\
    \hline
    Tweets2011 & PTHT & 0.05 & 0.10 & 0.11 & 0.12 & 0.12 \\
    \cline{2-7}
     & RL model (ours) & 0.36 & 0.39 & 0.38 & 0.35 & 0.31 \\
    \hline
    20 Newsgroups & PTHT & 0.13 & 0.13 & 0.13 & 0.13 & 0.12 \\
    \cline{2-7}
     & RL model (ours) & 0.35 & 0.30 & 0.27 & 0.25 & 0.22 \\
    \hline
  \end{tabular}
  \caption{NPMI coherence comparison between PTHT model and RL model for each number of topics}
  \label{tab:piaht_npmi}
\end{table*}

\begin{table}[ht!]
    \centering
    \begin{tabular}{ |c|c| }
        \hline
        Hyperparameter & Value(s) \\
        \Xhline{2\arrayrulewidth}
        Meta-seed & 4174224060 \\
        Num. Seeds & 30 \\
        Num. Epochs & 1000 \\
        Data Set & 20 Newsgroups \\
        Vocab Size & 2000 \\
        Embedding & SBERT \\
        Num. Topics ($N$) & 20 \\
        Inference Dropout & 0.2 \\
        Policy Dropout & 0.0 \\
        Inference Layers & [128, 128] \\
        Activation & GELU \\
        Initialization & $\bm{\rho} \sim N(0, 0.02)$ \\
        Normalization & Layer \\
        $\lambda$ & 5 \\
        Topic Words ($K$) & 10 \\
        RL policy & $\checkmark$ \\
        $\theta$ Softmax & $\times$ \\
        Learning Rate ($\alpha$) & 3e-4 \\
        Adam $\beta_{1}$, $\beta_{2}$ & 0.9, 0.999 \\
        Weight Decay & 0.01 \\
        Batch Size & 1024 \\
        Gradient Clipping & 1.0 \\
        \hline
    \end{tabular}
    \caption{Initial Experiments}
    \label{tab:hyp_initial}
\end{table}

\begin{table}[ht!]
    \centering
    \begin{tabular}{ |c|c| }
        \hline
        Hyperparameter & Value(s) \\
        \Xhline{2\arrayrulewidth}
        Meta-seed & 4174224060 \\
        Num. Seeds & 30 \\
        Data Set & 20 Newsgroups \\
        Embedding & \{BoW, SBERT\} \\
        $\theta$ / Policy Dropout & \{0.0, 0.2\} \\
        $\lambda$ & \{1, 5\} \\
        RL policy & \{$\checkmark$, $\times$\} \\
        $\theta$ Softmax & \{$\checkmark$, $\times$\} \\
        \hline
    \end{tabular}
    \caption{Ablation Study}
    \label{tab:hyp_ablation}
\end{table}

\subsection{Hyperparameters}
We show the hyperparameters for each experiment we performed. Experiment seeds are generated with a meta-seed for reproducibility. The meta-seed is randomly chosen from integers between 0 and $2^{32}$. Values in \{curly brackets\} indicate a search over multiple parameters. Values in [square brackets] indicate NN layer sizes (e.g. [128, 128] represents two layers of size 128).

\subsubsection{Initial Experiments and Ablation Study}
We use the same meta-seed for the ablation study as we did for the initial experiments. Hyperparameters for the initial experiments can be found in \autoref{tab:hyp_initial}. Further tables for all experiments will only show hyperparameters that differ from this table. Hyperparameters for the ablation study can be found in \autoref{tab:hyp_ablation}.

\subsubsection{Benchmarking Neural Topic Models}
We show hyperparameters for the comparison with \citet{doan2021benchmarking}. Hyperparameters for Snippets can be found in \autoref{tab:hyp_bntm_snippets}. 20 Newsgroups in \autoref{tab:hyp_bntm_20news}. W2E-title in \autoref{tab:hyp_bntm_w2etitle}. W2E-content in \autoref{tab:hyp_bntm_w2econtent}.

\subsubsection{Topic Modeling in Embedding Spaces}
Hyperparameters for the comparison with \citet{dieng2020topic} can be found in \autoref{tab:hyp_nyt}.

\subsubsection{Pre-training is a Hot Topic}
We show hyperparameters for the comparison with \citet{bianchi2020pre}. Data set and seed information can be found in \autoref{tab:hyp_ptht_seeds}. All other hyperparameters are the same for each data set; these can be found in \autoref{tab:hyp_ptht}.

\subsubsection{Contrastive Learning for NTM}
We show hyperparameters for the comparison with \citet{nguyen2021contrastive}. Some hyperparameters are already shown in \autoref{tab:hypsearch} and won't be shown again here. Data set and seed information can be found in \autoref{tab:hyp_contrastive_seeds}. Other hyperparameters are the same for each data set; these can be found in \autoref{tab:hyp_contrastive}. Hyperparameters for the policy dropout sweep can be found in \autoref{tab:hyp_contrastive_dropout}.

\subsection{Ablation Study}
We show full results from the ablation study in \autoref{tab:ablation_full}.

%% file: APPENDIX/parameters.tex
The number of parameters (P) in the model differs based on the total number of parameters across all inference layers (L), the number of topics (N), and the vocabulary size (V). Trainable parameters are the inference layers, the prior distribution of topics (N x 1), and the distribution of words over topics (V * N). Total parameters can be calculated with \autoref{eqn:parameters}.

\begin{equation}
    P = L + N + V * N
    \label{eqn:parameters}
\end{equation}

The largest model we use is for the Wikitext-103 data set with 200 topics. This model has 4,001,224 parameters.

%% file: APPENDIX/future_work.tex
We have identified some possible paths for future work. The SBERT embeddings could be fine-tuned during training rather than calculating them during pre-processing and freezing them during training. The RL formulation of our model could be extended to dynamic topic models \citep{blei2006dynamic}. More complex PG RL algorithms could be used rather than REINFORCE, or a baseline could be added to REINFORCE. Exploration techniques from RL could be applied. The influence of hyperparameters (e.g. inference network layer sizes) on varied corpora (e.g. those with large vocabularies) could be explored. The Laplace approximation of the Dirichlet prior could be replaced by a true Dirichlet prior, making use of the Dirichlet RT \citep{figurnov2018implicit} and a Dirichlet RL policy \citep{tian2022prescriptive}. Finally, $\lambda$ and the policy dropout could be scheduled during training to provide an automated tradeoff between topic diversity and coherence.

%% file: APPENDIX/final_tables.tex
\begin{table*}[ht]
    \centering
    \begin{tabular}{ |c|c|c|c| }
        \hline
        Data Set & Vocab Size & Meta-seed & Num. Seeds \\
        \Xhline{2\arrayrulewidth}
        Wiki20K & 2000 & 359491602 & 30 \\
        StackOverflow & 2236 & 1459046441 & 30 \\
        Google News & 8099 & 925040003 & 30 \\
        Tweets2011 & 5097 & 1321150024 & 30 \\
        20 Newsgroups & 2000 & 3277797161 & 30 \\
        \hline
    \end{tabular}
    \caption{PTHT Data Set Seeds}
    \label{tab:hyp_ptht_seeds}
\end{table*}

\begin{table*}[ht]
    \centering
    \begin{tabular}{ |c|c|c|c| }
        \hline
        Data Set & Vocab Size & Meta-seed & Num. Seeds \\
        \Xhline{2\arrayrulewidth}
        20 Newsgroups & 2000 & 1553571489 & 30 \\
        IMDb Movie Reviews & 5000 & 3747305026 & 30 \\
        Wikitext-103 & 20000 & 2672751736 & 30 \\
        \hline
    \end{tabular}
    \caption{CLNTM Data Set Seeds}
    \label{tab:hyp_contrastive_seeds}
\end{table*}

\begin{table}[ht!]
    \centering
    \begin{tabular}{ |c|c| }
        \hline
        Hyperparameter & Value(s) \\
        \Xhline{2\arrayrulewidth}
        Meta-seed & 193270011 \\
        Num. Seeds & 10 \\
        Data Set & Snippets \\
        Vocab Size & 4666 \\
        Num. Topics ($N$) & \{4, 8, 12, 16, 20, 24\} \\
        $\lambda$ & \{1, 3, 5, 10\} \\
        \hline
    \end{tabular}
    \caption{BNTM Snippets}
    \label{tab:hyp_bntm_snippets}
\end{table}

\begin{table}[ht!]
    \centering
    \begin{tabular}{ |c|c| }
        \hline
        Hyperparameter & Value(s) \\
        \Xhline{2\arrayrulewidth}
        Meta-seed & 1216545997 \\
        Num. Seeds & 10 \\
        Data Set & 20 Newsgroups \\
        Vocab Size & 4157 \\
        Num. Topics ($N$) & \{10, 20, 30, 40, 50, 60\} \\
        $\lambda$ & \{1, 3, 5, 10\} \\
        \hline
    \end{tabular}
    \caption{BNTM 20 Newsgroups}
    \label{tab:hyp_bntm_20news}
\end{table}

\begin{table}[ht!]
    \centering
    \begin{tabular}{ |c|c| }
        \hline
        Hyperparameter & Value(s) \\
        \Xhline{2\arrayrulewidth}
        Meta-seed & 4014169843 \\
        Num. Seeds & 10 \\
        Data Set & W2E-title \\
        Vocab Size & 3703 \\
        Num. Topics ($N$) & \{15, 30, 45, 60, 75, 90\} \\
        $\lambda$ & \{1, 3, 5, 10\} \\
        \hline
    \end{tabular}
    \caption{BNTM W2E-title}
    \label{tab:hyp_bntm_w2etitle}
\end{table}

\begin{table}[ht!]
    \centering
    \begin{tabular}{ |c|c| }
        \hline
        Hyperparameter & Value(s) \\
        \Xhline{2\arrayrulewidth}
        Num. Topics ($N$) & \{25, 50, 75, 100, 150\} \\
        $\lambda$ & 1 \\
        \hline
    \end{tabular}
    \caption{Pre-training is a Hot Topic}
    \label{tab:hyp_ptht}
\end{table}

\begin{table}[ht!]
    \centering
    \begin{tabular}{ |c|c| }
        \hline
        Hyperparameter & Value(s) \\
        \Xhline{2\arrayrulewidth}
        Meta-seed & 1359128464 \\
        Num. Seeds & 10 \\
        Data Set & W2E-content \\
        Vocab Size & 10508 \\
        Num. Topics ($N$) & \{15, 30, 45, 60, 75, 90\} \\
        $\lambda$ & \{1, 3, 5, 10\} \\
        \hline
    \end{tabular}
    \caption{BNTM W2E-content}
    \label{tab:hyp_bntm_w2econtent}
\end{table}

\begin{table}[ht!]
    \centering
    \begin{tabular}{ |c|c| }
        \hline
        Hyperparameter & Value(s) \\
        \Xhline{2\arrayrulewidth}
        Meta-seed & 2337766308 \\
        Num. Seeds & 1 \\
        Num. Epochs & 20 \\
        Data Set & New York Times \\
        Vocab Size & 10283 \\
        Num. Topics ($N$) & 300 \\
        Inference Dropout & 0.5 \\
        Inference Layers & [512, 512] \\
        $\lambda$ & 1 \\
        Topic Words ($K$) & 10* \\
        Batch Size & 32768 \\
        \hline
    \end{tabular}
    \caption{Topic Modeling in Embedding Spaces (*We use $K=25$ to calculate topic diversity for the final model.)}
    \label{tab:hyp_nyt}
\end{table}

\begin{table}[ht!]
    \centering
    \begin{tabular}{ |c|c| }
        \hline
        Hyperparameter & Value(s) \\
        \Xhline{2\arrayrulewidth}
        Num. Epochs & 2000 \\
        Num. Topics ($N$) & \{50, 200\} \\
        \hline
    \end{tabular}
    \caption{Contrastive Learning for NTM}
    \label{tab:hyp_contrastive}
\end{table}

\begin{table*}[ht]
    \centering
    \begin{tabular}{ |c|c| }
        \hline
        Hyperparameter & Value(s) \\
        \Xhline{2\arrayrulewidth}
        Meta-seed & 3432645033 \\
        Num. Seeds & 30 \\
        Data Set & 20 Newsgroups \\
        Num. Epochs & 2000 \\
        Num. Topics ($N$) & 50 \\
        Inference Dropout & 0.5 \\
        Policy Dropout & \{0.0, 0.1, 0.2, 0.3, 0.4, 0.5, 0.6, 0.7, 0.8, 0.9\} \\
        Inference Layers & [128, 128] \\
        $\lambda$ & 1 \\
        \hline
    \end{tabular}
    \caption{CLNTM Dropout Sweep}
    \label{tab:hyp_contrastive_dropout}
\end{table*}

\begin{table*}[ht!]
    \centering
    \begin{tabular}{ |c|c|C{0.05\textwidth}|c|c||c|c| }
        \hline
        RL Policy & Embedding & $\lambda$ & $\theta$ Softmax & $\theta$ / Policy Dropout & Coherence & Diversity \\
        \Xhline{2\arrayrulewidth}
        $\times$ & BoW & 1 & $\checkmark$ & 0.0 & 0.2906 & 0.8457 \\
        $\times$ & BoW & 1 & $\times$ & 0.0 & 0.2373 & 0.6943 \\
        $\checkmark$ & BoW & 1 & $\checkmark$ & 0.0 & 0.2748 & 0.8905 \\
        $\checkmark$ & BoW & 1 & $\times$ & 0.0 & 0.2738 & 0.8707 \\
        $\times$ & BoW & 5 & $\checkmark$ & 0.0 & 0.2526 & 0.6598 \\
        $\times$ & BoW & 5 & $\times$ & 0.0 & 0.2619 & 0.6928 \\
        $\checkmark$ & BoW & 5 & $\checkmark$ & 0.0 & 0.2032 & 0.5965 \\
        $\checkmark$ & BoW & 5 & $\times$ & 0.0 & 0.3379 & 0.9403 \\
        $\times$ & BoW & 1 & $\checkmark$ & 0.2 & 0.2650 & 0.7390 \\
        $\times$ & BoW & 1 & $\times$ & 0.2 & 0.2193 & 0.5195 \\
        $\checkmark$ & BoW & 1 & $\checkmark$ & 0.2 & 0.2082 & 0.5692 \\
        $\checkmark$ & BoW & 1 & $\times$ & 0.2 & 0.2798 & 0.7740 \\
        $\times$ & BoW & 5 & $\checkmark$ & 0.2 & 0.2526 & 0.6222 \\
        $\times$ & BoW & 5 & $\times$ & 0.2 & 0.2257 & 0.5768 \\
        $\checkmark$ & BoW & 5 & $\checkmark$ & 0.2 & 0.1222 & 0.314 \\
        $\checkmark$ & BoW & 5 & $\times$ & 0.2 & 0.3284 & 0.8092 \\
        $\times$ & SBERT & 1 & $\checkmark$ & 0.0 & 0.2845 & 0.6207 \\
        $\times$ & SBERT & 1 & $\times$ & 0.0 & 0.2948 & 0.5995 \\
        $\checkmark$ & SBERT & 1 & $\checkmark$ & 0.0 & 0.2158 & 0.8080 \\
        $\checkmark$ & SBERT & 1 & $\times$ & 0.0 & 0.3414 & 0.9070 \\
        $\times$ & SBERT & 5 & $\checkmark$ & 0.0 & 0.2726 & 0.4458 \\
        $\times$ & SBERT & 5 & $\times$ & 0.0 & 0.2795 & 0.4530 \\
        $\checkmark$ & SBERT & 5 & $\checkmark$ & 0.0 & 0.1932 & 0.6927 \\
        $\checkmark$ & SBERT & 5 & $\times$ & 0.0 & $\textbf{0.3848}$ & $\textbf{0.9530}$ \\
        $\times$ & SBERT & 1 & $\checkmark$ & 0.2 & 0.2532 & 0.6063 \\
        $\times$ & SBERT & 1 & $\times$ & 0.2 & 0.2554 & 0.5430 \\
        $\checkmark$ & SBERT & 1 & $\checkmark$ & 0.2 & 0.1133 & 0.5520 \\
        $\checkmark$ & SBERT & 1 & $\times$ & 0.2 & 0.3649 & 0.7663 \\
        $\times$ & SBERT & 5 & $\checkmark$ & 0.2 & 0.2435 & 0.4478 \\
        $\times$ & SBERT & 5 & $\times$ & 0.2 & 0.2080 & 0.3698 \\
        $\checkmark$ & SBERT & 5 & $\checkmark$ & 0.2 & 0.0967 & 0.9227 \\
        $\checkmark$ & SBERT & 5 & $\times$ & 0.2 & 0.3769 & 0.7315 \\
        \hline
    \end{tabular}
    \caption{Full Results from Ablation Study}
    \label{tab:ablation_full}
\end{table*}